\documentclass[letterpaper, 10 pt, conference]{ieeeconf}  

\IEEEoverridecommandlockouts                              
\usepackage{times}
\usepackage{graphicx}

\DeclareGraphicsExtensions{.png,.pdf,.jpg,.eps}

\usepackage{subfigure}
\usepackage{amssymb}
\usepackage{latexsym}
\usepackage{amsmath}

\newcommand{\ex}{\mathbb{E}}

\newcommand{\T}{\mathcal{T}}

\newcommand{\Hm}{\mathbf{H}}
\newcommand{\I}{\mathbf{I}}
\newcommand{\K}{\mathbf{K}}

\newcommand{\Pm}{\mathbf{P}}
\newcommand{\Q}{\mathbf{Q}}
\newcommand{\R}{\mathbf{R}}

\newcommand{\comment}[1]{}

\newcommand{\x}{{\mathbf{x}}}
\newcommand{\y}{\mathbf{y}}

\newcommand{\vt}{\mathbf{v}_t}

\newcommand{\thetav}{\mathbf{\theta}}

\newcommand{\N}{\mathcal{N}}

\newcommand{\beq}{\begin{equation}}
\newcommand{\eeq}{\end{equation}}
\newcommand{\ba}{\begin{array}}
\newcommand{\ea}{\end{array}}

\newcounter {examplecounter}[section]

\newcounter {algorithmcounter}[section]

\title{\LARGE \bf
Body Schema Acquisition through Active Learning
}

\author{Ruben Martinez-Cantin, Manuel Lopes and Luis Montesano%
\thanks{R. Martinez-Cantin is with the Institute of Systems and Robotics, Instituto Superior T\'ecnico
{\tt\small rmcantin@isr.ist.utl.pt}}%
\thanks{M. Lopes is with the Centre for Robotics and Neural Systems of the University of Plymouth
{\tt\small manuel.lopes@plymouth.ac.uk}}%
\thanks{L. Montesano is with the Departamento de Informatica e Ingenieria de Sistemas. Universidad de Zaragoza
{\tt\small lmontesa@unizar.es}}}

\begin{document}

\maketitle
\thispagestyle{empty}
\pagestyle{empty}

\begin{abstract}
We present an  active learning algorithm for the problem of body schema learning, i.e. estimating a kinematic model of a serial robot. The learning process is done online using \emph{Recursive Least
Squares} (RLS) estimation, which outperforms gradient methods usually applied in the literature. In addiction, the method provides the required information to apply an active learning algorithm to find the optimal set of robot configurations and observations to improve the learning process. By selecting the most informative observations, the proposed method minimizes the required amount of data.

We have developed an efficient version of the active learning algorithm to select the points in real-time. The algorithms have been tested and compared using both simulated environments and a real humanoid robot.
\end{abstract}

\section{Introduction}

In this paper we study the problem of efficiently learning a physical model of a (humanoid) robot. Knowing the forward and inverse kinematic functions of a serial manipulator, or humanoid robot, is the fundamental stone where the rest of motion capabilities are built on. Many approaches address this problem with non-parametric methods \cite{aaronetall01learninvkin,macl07manifold,Nguyen-Tuong2008} without relying on the actual geometry. Using physical (geometrical) models to learn the body schema of kinematic chains, besides providing directly the forward and inverse kinematics, has the advantage of getting a robust model, which is usually more general and serves as the base of most robot control methods. 

This problem was first studied for industrial robots where several authors proposed methods for precise robot calibration based on least squares (LS) estimation \cite{bennett88robotcalibration,maric99learndh}. These methods rely on a dataset of joint and end-effector positions. The end-effector is tracked using high accuracy external sensors, such as lasers and magnetic sensors, which provide high resolution observations along the whole configuration space.

More interesting, though, is the work from the epigenetic robotics community, where biologically inspired setups are used to study the learning process in humans. Recent experiments on humans and monkeys have shown that we learn our own body schema combining proprioceptive and visual information \cite{Maravita04TCS}. The learning process is done online, as new data arrive, being able to adapt to new configurations.
Based on this biological concept, Hersch et al. \cite{micha08bodyschema} suggest an online parametric learning of the body schema, based on gradient descent for Stochastic Approximation. They also present results of the method with immediate body changes. In \cite{burgard08bodyschema}, the authors address the problem of learning the structure of an arm when the sequence and number of links is unknown. The learned model allows the robot to follow trajectories and adapt to large modifications of the kinematics, e.g. stuck and deformed joints. However, the method requires observations along the whole robot arm to disambiguate between models, for what it relies on external sensors. For robots consisting of several kinematic chains, evolutionary algorithms were used to learn and adapt controllers for locomotion \cite{bongard06selfmodeling}.

\begin{figure}
\centering
\subfigure[]{
\includegraphics[width=0.40\columnwidth]{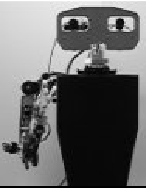}
\label{fig:balta}}
\subfigure[]{
\includegraphics[width=0.40\columnwidth]{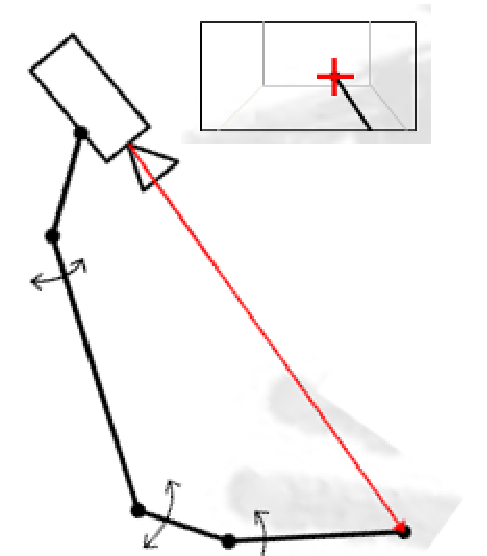}
\label{fig:schema}}
\caption{(a) Humanoid robot with a binocular head and a 6 dof arm and (b) sketch of the body schema learning problem. }
\end{figure}

A common drawback of all the previous methods is the inefficient exploration of the robot workspace. Typically, they use random perturbations to explore the configuration space of the robot, since finding a systematic way to explore all the degrees of freedom taking into account the joint limits is not an easy task \cite{Choset01AMAI}. In this paper, we suggest a new method based on active learning which explores the configuration space searching for the most informative measurements. A similar approach is used in \cite{brock08bodyschema} to learn generic planar kinematics of tools and articulated objects, using a reinforcement learning model based on the number of articulations discovered. In our case, the active selection is based on an Bayesian experimental design criterion \cite{Chaloner:1995} that provides a cost function in terms of information gain for each potential measurement. The optimization of the cost function is solved using the DIRECT algorithm \cite{Jones:1993}, an efficient global optimization algorithm, to select the measurement with the lowest expected cost. 
Since the proposed experimental design criterion requires the estimation of the posterior uncertainty of the parameters, we use a \emph{Recursive Least Squares} (RLS) estimator instead of stochastic gradient widely used in the literature.

In summary, the main contributions of this paper are: (i) a more efficient learning method based on RLS for estimating the body schema parameters , and (ii) an active learning strategy that uses the posterior uncertainty provided by the RLS to significantly reduce the number of motions required to obtain a good estimate. 
Both contributions are evaluated and compared to a gradient based methods based on \cite{micha08bodyschema}. The evaluation has been done using simulated data as well as a real humanoid robot with a 6 degrees of freedom arm (see Fig. \ref{fig:balta}). The parameters of the body schema are learnt based on the joint space configuration and the sensor readings of the head camera tracking the end-effector. The results show that the error and the number of observations required to converge are both reduced compared to prior work.
The key of the improvement is the synergy of two effects. First, the use of the RLS method provides a faster convergence compared to the gradient descend method, because RLS uses the uncertainty to adapt every optimization step. Second, the active learning procedure provides a better exploration of the configuration space by selecting the most informative measurements according to the parameters uncertainty.

The remainder of the paper is organized as follows. Section \ref{sec:kalman} formalizes the body schema learning as a parameter estimation problem and introduces the Recursive Least Squares estimator. Section \ref{sec:activelearning} shows how to perform active learning by selecting an exploration policy based on confidence information of the previous estimator. In Section \ref{sec:results}, we present the experimental results and compare our method to previous state of the art methods. We conclude with a discussion in Section \ref{sec:conclusions}.

\section{Recursive Least Squares for Body Schema Learning}
\label{sec:kalman}

Online learning of the body schema can be formalized as a sequential parameter estimation problem. Let $\x$ be a set of parameters that represents the robot schema. For a serial manipulator, $\x = ({\x}_1,\cdots,{\x}_n)$ where each ${\x}_i$ represents the link size, localization and orientation respective to the actuated joint and $n$ is the number of degrees of freedom. 
Our objective is to estimate these parameters by positioning the manipulator at different configurations (joint angles of the degrees of freedom) ${\thetav}_k$ and using an external sensor to obtain measurements of the arm. 
For a serial manipulator, the position of the joints and the end effector contains information about the kinematic chain. 
Let ${\y}_k$ be the measurement of the end-effector for a given configuration $\thetav_k$. The relation of those two variables with the parameters ${\x}$ is given by the observation model,
\begin{equation}
 \label{eq:obsmodel}
 \y_t = h(\x,\thetav_t) + \vt,
\end{equation}
where $\vt$ is a random noise characterized by a zero-mean normal distribution $\vt \sim \N(0, \R)$ that defines a likelihood model $p(\y_{t} \mid \x , \thetav_t)$. Note that the observation model implicitly considers the kinematic model, e.g.: the robot arm kinematic chain; and the sensor model, e.g.: the camera projection function. The objective is to estimate the parameters ${\x}$ based on a set of configurations ${\thetav}_{1:t}$ and sensor measurements ${\y}_{1:t}$. Figure \ref{fig:schema} illustrates the estimation problem in a schematic way. For simplicity, the camera parameters are not included in the formulation, that is, they are assumed to be known in advance. However, one could also estimate them along with the kinematic parameters.

We use a Bayesian approach to compute the posterior distribution $p({\x} | {\y}_{1:t}, {\thetav}_{1:t})$ and, based on the latter, define a selection criterion to find the most informative configurations (see Section \ref{sec:activelearning}). 
In general, Bayesian inference is only tractable in few cases, such as linear systems. In our case, the body schema learning problem is nonlinear due to the measurement model $h(\cdot)$. Therefore, we use a linearized version of the model at each configuration and apply the RLS method where the prior, posterior and likelihood functions are modeled as Gaussian distributions, $p({\x} | {\y}_{1:t}, {\thetav}_{1:t}) \simeq \N(\hat{\x}_t, \Pm_t)$. Under these conditions, the RLS estimate can be computed recursively in closed form. This method has shown very good results in practice for parameter estimation and filtering, being the later the well-known extended Kalman filter (EKF).

The RLS algorithm can be summarized in two update equations for the estimate and
its covariance:
\begin{align}
 \hat{\x}_t &= \hat{\x}_{t-1} + \K_t \left(\y_t - \Hm_t \x_{t-1}  \right) 
\label{eq:updatex} \\
 \Pm_t &= (\I-\K_t \Hm_t) \Pm_t (\I-\K_t \Hm_t)^T + \K_t \R_t \K_t^T
\label{eq:updatep}
\end{align}
where $\Hm_t$ is the Jacobian of the observation model for a given configuration $h(\cdot,\thetav_t)$ 
and the gain factor $\K_t$ is 
\begin{equation}
 \K_t = \Pm_t \Hm^T_t (\Hm \Pm_t \Hm^T_t + \R_t).
\end{equation}

For parameter estimation, linearization errors can be compensated by injecting some extra noise in the parameter distribution after certain number of iterations. This term is typically called \emph{stabilizing noise} in filtering theory or \emph{nugget} in the regression literature:
\begin{equation}
 \Pm_t = \Pm_t + \sigma^2_{n}\I
\end{equation}
In addition, this extra noise may also provide some degree of adaptability, similar to the body schema learning in humans. The drawback is that we limit the accuracy of the estimator that will suffer small oscillations in the stationary regime.

\section{Active Learning}
\label{sec:activelearning}
From the previous Section, we can compute the posterior distribution of the parameters as new pairs of configurations and measurements arrive. We now study active learning strategies to select the most informative configurations and reduce the number of measurements required to estimate the parameters ${\x}$. For this purpose, we follow the standard procedure in active learning which consists in defining an appropriate cost function and, then, look for the  measurement that minimizes it. The new measurement ${\y_t}$ is used to update the posterior of ${\x}$. Then, the active learning process is repeated using the new posterior distribution. 

\subsection{Cost function}
\label{sec:cost}
Intuitively, we want to select the next $\thetav$ whose observation will most reduce the (expected) error in the estimate.
The Bayesian experimental design literature \cite{Chaloner:1995} has studied several optimality criteria to achieve this objective. Based on the results of \cite{Sim:2005}, we use the A-optimality criterion and define the cost function as the expected mean squared error of the robot parameters.  Given our  linear-Gaussian approximation of the posterior, the cost function is given by the expected trace of the covariance matrix \cite{Sim:2005}:
\begin{align}
C(\thetav_T)&=\ex_{p(\x,\y_{\T}|\y_{1:T-1},\thetav_{1:T})} \left[(\widehat{\x}_\T-\x)^T
(\widehat{\x}_\T-\x) \right]\\
 &\simeq \ex_{p(\x,\y_{\T}|\y_{1:T-1},\thetav_{1:T})} tr(\Pm_T)
\label{eqn:amsecost}
\end{align}
where $\widehat{\x}_T = \ex_{p(\x|\y_{1:T},\thetav_{1:T})}[\x]$ is the posterior estimate of the state at time $T$ and $\Pm_T$ is the corresponding covariance matrix. 
Note that time index $T$ is in the future, since we want to select new configurations $\thetav_T$ and obtain the corresponding measurements $\y_T$. Although in the general case, we could plan several configurations ahead, for simplicity, let us consider one time step in the future. In that case, we already have available measurements up to $T-1$ and the corresponding posterior $p(\x \mid \y_{1:\T-1},\thetav_{1:T-1})$. 

The expectation in (\ref{eqn:amsecost}) is computed based on simulating future observations \cite{MartinezCantin09AR,Ng:2000,MartinezCantin07RSS}. To simulate the measurements, we use the parameters' posterior and propagate it through the observation model given a configuration $\theta$. In principle, one could use Monte Carlo techniques to approximate the posterior and evaluate the cost \cite{Ng:2000, MartinezCantin07RSS}. However, since the observation model is linearized and the posterior distribution is Gaussian, we use the approximation proposed in \cite{MartinezCantin09AR}. The main idea is to use the current estimate $\widehat{\x}_{T-1}$ to compute the {\em maximum a posteriori} of future observations. This procedure is actually deterministic under linear gaussian models. It has been shown experimentally that it outperforms Monte-Carlo simulations when the Gaussian approximation is good enough. 
The most informative measurement corresponds to the optimal configuration i.e., the one that minimizes the cost function $\thetav^* = \arg \min_\thetav C(\thetav)$.
Figure \ref{fig:approach} summarizes the steps to simulate observations.

\begin{figure}
\begin{center}
\par \hspace{-.5mm} \framebox[0.98\linewidth][l]{
\begin{minipage}[l]{0.93\linewidth}\vspace{2mm}
 {\footnotesize

\begin{enumerate}
\item For $j=1:MaxNumberOfIterations$
\begin{enumerate}
\item Simulate the next robot configuration $\thetav_j$.
\item Simulate the new observation based on the current estimate and the
observation model $\y_t^{(i)} = h(\x_{t}^{(i)},\thetav_j)$.
\item Generate a simulated posterior distribution using the current posterior distribution as a prior, the new observations and Eqs. (\ref{eq:updatex}) and (\ref{eq:updatep}).
\item Evaluate the approximate cost function (\ref{eqn:amsecost}) based on the simulated posterior $C(\thetav_j)$.
\item Generate a new sample configuration $\thetav_{j+1}$ following Section \ref{sec:opt}.
\begin{enumerate}
	\item Find the optimal hypercubes according to Eqs. \eqref{eq:testdirect}.
	\item Divide the hypercubes.
	\item $\thetav_{j+1}$ is the center of the new hypercubes.
\end{enumerate}
\end{enumerate}
\item Choose $\thetav^* = \arg \min_\thetav C(\thetav)$.
\end{enumerate}
 \vspace{2mm} \small}
\end{minipage}}
\end{center}
\caption{Steps for the active selection of the most informative configurations.}
\label{fig:approach}
\end{figure}

\subsection{Action selection through efficient optimization}
\label{sec:opt}

Having restricted the problem of selecting the most informative robot configurations to one of finding the global minimum of a cost function, we may use many of the existing techniques for efficient global optimization.
In our active learning setting, we have a computational budget for optimization. That is, the overall cost cannot exceed the cost of processing large datasets of random data or, even worse, the time to sweep the whole space with the robot. Furthermore, we would like to achieve online learning, where the action selection can be done in real-time. Since the evaluation of the cost function is expensive, the number of iterations and function evaluations should remain small.

In our setup, all the functions involved are continuous in a bounded set defined by the camera field of view and joint limits. Since action and parameter spaces are large-dimensional and continuous, it is not possible to use methods based on discretization \cite{Smallwood:1973}. To optimize the cost function, we use the DIRECT method \cite{Jones:1993}. This algorithm is a particular instance of Lipschitzian optimization.

Lipschitzian optimization is a simple but powerfull method for global optimization that relies on a deterministic search in a multidimensional hypercube. Any Lipschitzian algorithm assumes that the cost function satisfies
\begin{equation}
 |f(\thetav) - f(\thetav')| \leqslant K|\thetav-\thetav'| \; \;\;\; \forall \thetav, \thetav' \in S
\end{equation}
for a positive constant $K$ and any closed set $S$. This assumption allows us to set a lower bound for the function between any two known points in a closed set. The lower bound is used to evaluate the function, divide the set in smaller sets and refine the search recursively. The algorithm presented in \cite{Jones:1993} (called DIRECT, for \emph{divide rectangles}) uses hyperrectangles as closed sets. The center point of each set is used for evaluation. 

The advantage of using the central point is twofold.
First, it requires to evaluate only one point for each hyperrectangle despite the number of dimensions, contrary to the $2^n$ points required to evaluated the corners. In theory, as any global black-box optimization algorithm, it has an exponential worse-case scenario; but that condition is never met, under mild conditions. Thus, the central point sampling allows to work in high-dimensional spaces. According to the authors, the algorithm is efficient for 10-20 dimensions, which is more than enough for our problem.

\begin{figure}
\begin{center}
\subfigure[]{\includegraphics[width=0.25\columnwidth]{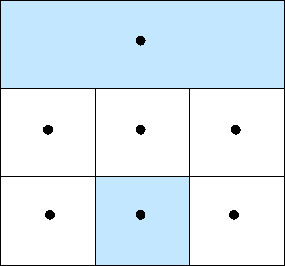}}
\subfigure[]{\includegraphics[width=0.25\columnwidth]{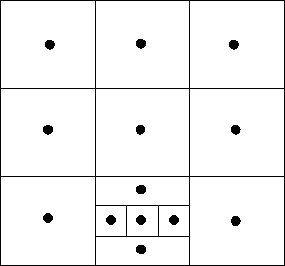}}
\subfigure[]{\includegraphics[width=0.40\columnwidth]{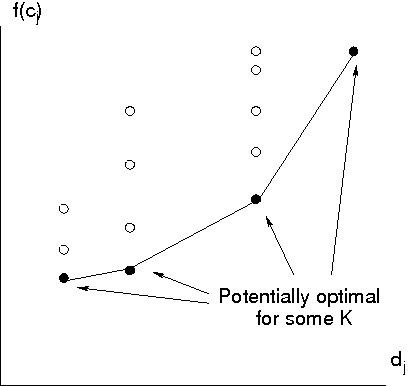}}
\caption{Example of an iteration of DIRECT for a 2D function. The division from the previous iteration is shown in a). We evaluate all the middle points and find the potentially optimal hypercubes (shaded). Then, we subdivide those hypercubes as shown in b) and iterate. The potentially optimal hypercubes are the points that fall in the lower convex hull shown in c), where $f(c_j)$ is the value of the function in the central point, and $d_j$ is the distance of the central point to the extreme. Adapted from \cite{Jones:1993}.}
\label{fig:direct}
\end{center}
\end{figure}

Second, the center point samples allow to compare the lower bound of each subset for any Lipschitz constant $K$. Therefore, we can compare the potential lower bound of each hyperrectangle without knowing $K$. Once we have the lower bound of each hyperrectangle, we can identify potentially optimal hyperrectangles $j \in [1 \cdots m]$ that satisfy
\begin{align}
 \exists \tilde{K} > 0 \mid f(c_j) - \tilde{K}d_j &\leqslant f(c_i) - \tilde{K}d_i  \; \;\;\; \forall i= 1 \cdots m \label{eq:testdirect}\\
f(c_j) - \tilde{K}d_j &\leqslant f_{min} - \epsilon |f_{min}|
\end{align}
where $c_i$ is the central point of hyperrectangle $i$ and $f_{min}$ is the minimum value found so far. The first condition selects all potentially optimal points for any positive $\tilde{K}$ by just selecting the piecewise linear function that connects the points with minimum $f(c)$ for any $d_j$. Figure \ref{fig:direct} provides a simple intuition of the method. 
The second condition guarantees that at any iteration, there is a non-trivial improvement in order to guarantee global convergence and to avoid being stuck in flat local minima. In our robotics setup, this condition is specially critical, since small improvements in the cost functions are pointless in terms of learning, while global exploration is fundamental to find good observations. Nevertheless, the performance of the method is good for any reasonable value of $\epsilon$, provided that the dataset is normalized to the unit hypercube beforehand \cite{Jones:1993}.

As we require to set a budget for the optimization process, it seems more suitable to use the extension of the DIRECT algorithm, called DIRECT-l \cite{Gablonsky:2001}. DIRECT-l reduces the computational cost of every sweep over the function by clustering the rectangles. The improvement in time comes at the cost of biasing the optimization towards local search solutions, although maintaining global convergence.

\section{Results}
\label{sec:results}
In this section we evaluate the proposed method. We evaluate the improvement of the active learning strategy over a passive RLS estimator and a passive gradient descent \cite{micha08bodyschema}. We start by presenting the real robotic platform used in the experiments and the kinematic representation chosen. Then, we present some simulation results to analyze our algorithm in terms of reliability, precision and convergence speed, Finally, we illustrate the advantages of our active learning method in the real robot.
 
\subsection{Setup}
\label{sec:setup}
We use Baltazar \cite{macl04baltazar} (see Fig. \ref{fig:balta}), a humanoid-torso consisting of a six degrees-of-freedom (dof) arm and a binocular
head with four dof. For the experiments we assume that the head kinematics is known. A fiducial marker is attached to the end of the arm. By using calibrated cameras and the ARToolKit tracker \cite{ARToolkit}, we obtain \textit{3D} observations of the end-effector position. The use of orientation information could be easily accommodated and would provide extra information. These are the only measurements used during learning that, under certain conditions, are enough
for robot identification \cite{Khalil02robotidentification}. 

After calibrating the cameras, we performed a set of test measurements and compute the observation error. We found it to be around $\pm 5 cm$ per dimension in $90\%$ of the cases. The cameras are equipped with large field-of-view lenses. Even so, as the camera-arm distance is small, the end-effector goes out of the image. This happens with amplitudes of $\pm 20\deg$ around an initial position in the center of the image. To alleviate this problem, the head performs servoing to keep the marker inside the image. However, the visible space is still very constrained with respect to the actual configuration space of the arm, making more difficult to find informative robot configurations.

\subsection{Kinematic representation}
\label{sec:kinematics}

We follow the twist representation as presented by
\cite{robmanipulation94sastry}. It has several advantages when compared with the
more usual Denavit-Hartenberg parameterization. First, it needs only the
specification of the base and end-effector frames of reference and can represent
any kind of serial robot. Also the geometrical meaning of the parameters is more
intuitive, i.e. the parameters represent directly the axis of rotation ${\bf w}$
and an arbitrary point in this axis ${\bf v}$.

The direct kinematics of a serial manipulator has the following expression

\[
{^0_n}T({\thetav}) = \prod_{i=1\ldots n} e^{\xi \theta_i} {^0_n}T({\thetav}=0)
\]

where ${^0_n}T({\thetav})$ represents the relation between the end-effector
frame and the base frame for a given configuration ${\thetav}$ of the robot. Each axis is
represented by the exponential map that maps points from an initial position to
the transformed position using the twist parameters $\xi = \left[\mathbf{v}, \mathbf{w} \right]^T$. That is,
\begin{equation}
 e^{\xi {\thetav}} = 
\left[\begin{array}{cc}
	e^{{\bf w} {\thetav}} & (I - e^{{\bf w} {\thetav}}) ({\bf w} \times {\bf
v}) + {\bf w} {\bf w}^T {\bf v} {\thetav} \\
	\mathbf{0}	& 1
\end{array}\right].
\end{equation}

Note that each transformation is not a map between different coordinate frames,
but the transformation that the points undergo with a given motion in the base
frame of reference.

\begin{figure*}
 \centering
\subfigure[]{\includegraphics[width=0.6\columnwidth]{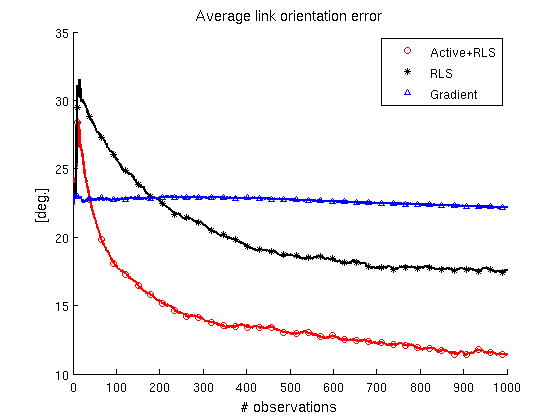}\label{fig:res6dofplanar:ang}}
\subfigure[]{\includegraphics[width=0.6\columnwidth]{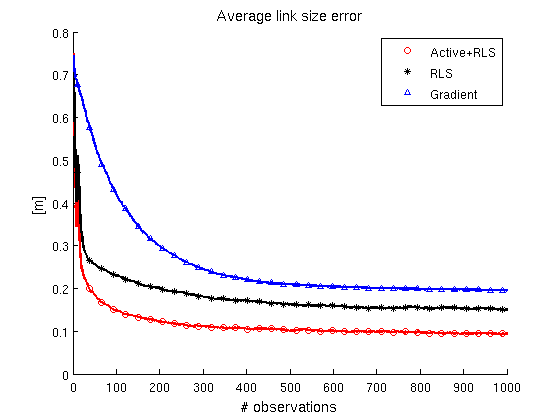}\label{fig:res6dofplanar:l}}
\subfigure[]{\includegraphics[width=0.6\columnwidth]{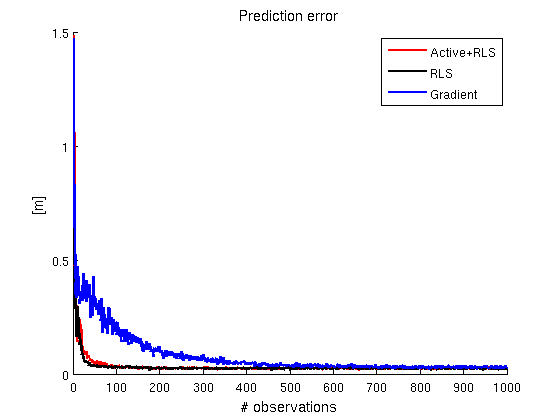}\label{fig:res6dofplanar:pred}}
\caption{Convergence of the body schema learning for a $6$ dof robot. (a) mean error in the relative orientations of the joint angles, (b)
mean error in the location of the joint angle and (c) prediction error during
the learning. Each figure shows the results for random exploration using the RLS, the active approach and an stochastic gradient method.}
\label{fig:res6dofplanar}
\end{figure*}

\begin{figure*}
 \centering
\subfigure[]{ \includegraphics[width=0.6\columnwidth]{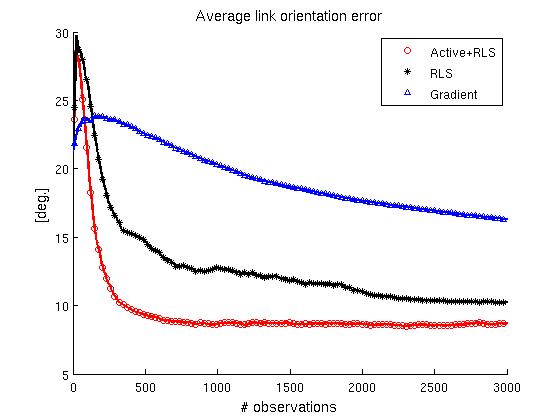}\label{fig:res12dofplanar:ang}}
\subfigure[]{\includegraphics[width=0.6\columnwidth]{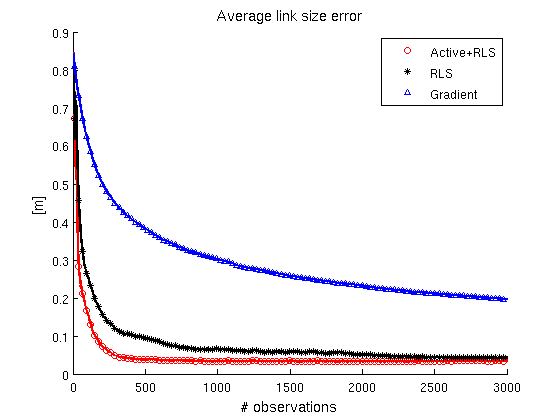}\label{fig:res12dofplanar:l}}
\subfigure[]{\includegraphics[width=0.6\columnwidth]{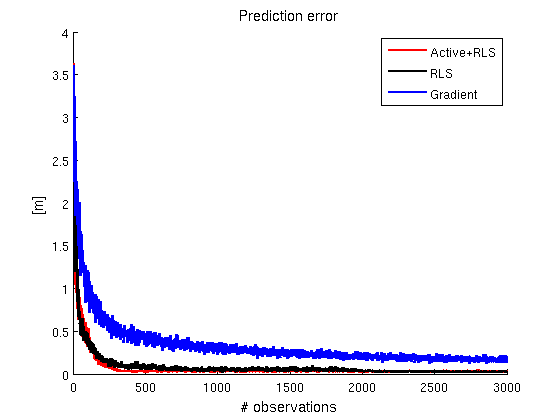}\label{fig:res12dofplanar:pred}}
\caption{Convergence of the body schema learning for the simulated robot with
$12$ dof. (a) mean error in the relative orientations of the joint angles, (b)
mean error in the location of the joint angle and (c) prediction error during
the learning. Each figure shows the results for random exploration using the RLS, the active approach and an stochastic gradient method.}
\label{fig:res12dofsimul}
\end{figure*}

\subsection{Simulations}
\label{sec:simulations}

The first test was done on a simulated version of Baltazar. Figure \ref{fig:res6dofplanar} shows the result of 20 runs of the algorithms
with a random initialization of the parameters in a uniform hyper-cube. Each experiment consisted of 1000 observations of the end-effector from different robot configurations. We provide results for the RLS based parameter estimation with random motions and for the corresponding active learning action selection according to the method presented in Section \ref{sec:activelearning}. The orientation error is computed as the average of the errors of the orientation of every joint with respect to the orientation of every other joint. For comparison purposes we also implemented the stochastic gradient presented in \cite{micha08bodyschema}. The observation noise used during the simulations was $\mathbf{v} \sim \mathcal{N}(0,\sigma^2_R \I)$ with $\sigma^2_R=0.0001$. 

The first observation is that all methods rapidly reduce the prediction error (see Fig. \ref{fig:res6dofplanar:pred}). Actually, for this
robot structure, they all require few observations. However, a small prediction error does not guarantee an unbiased convergence of robot parameters. This is because the limited exploration volume and the high number of parameters are prone to overfit. This behavior is expected since the exploration volume directly impacts the observability of the system. 

Figures \ref{fig:res6dofplanar:ang} and \ref{fig:res6dofplanar:l} show the robot parameters convergence. In average, both methods quickly learn the parameters, specially the link size. The active learning method improved this result by achieving better and faster results. An inspection of the individual test revealed that this was partly due to the fact that random methods sometimes get trapped in local minima.

In order to study the scalability of the system, the second set of experiments were performed on a simulated model of a $12$ dof humanoid robot arm. Figure \ref{fig:res12dofsimul} summarizes the results. The main difference is that the gradient converges very slowly compared to the RLS method. An analysis of each link separately revealed that, for both the $6$ and $12$ dof cases, the gradient reduces the prediction error mainly by correcting the first and last link of the kinematic chain. Thus, the prediction error decreases rapidly. However, as the number of dof increase, there are more parameters to estimate and the algorithm takes longer to correct the intermediate joints. 
For the $12$ dof case, the locations of the links are learnt much faster than their orientations. This affects specially the random sampling strategy which is highly inefficient compared to the active one. By inspecting the joints separately, we found out that the last two degrees of freedom, those corresponding to the wrist, are more difficult to estimate correctly. This is because their motion does not modify the observation as much as the other joints and therefore, updates are slower. That is, the signal-noise ratio of the wrist movements is very small or close to 1.

Regarding computational time, the 12 dof robot takes around 0.3 seconds to select the next optimal configuration according to criterion \ref{eqn:amsecost}, which is close to real time planning. Finally, we also compared the results with the Bayesian optimization method presented in \cite{MartinezCantin07RSS} for active learning. However, the performance increase was not enough to compensate the additional computational cost.

\begin{figure*}
	\centering
\subfigure[]{\includegraphics[width=0.6\columnwidth]{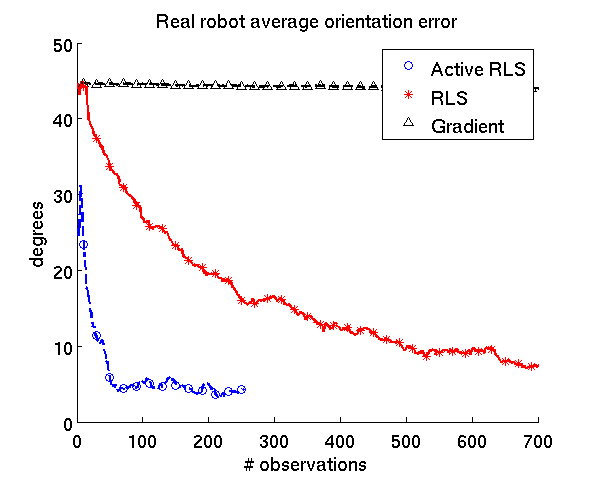}\label{fig:resbaltazar:ang}}
\subfigure[]{\includegraphics[width=0.6\columnwidth]{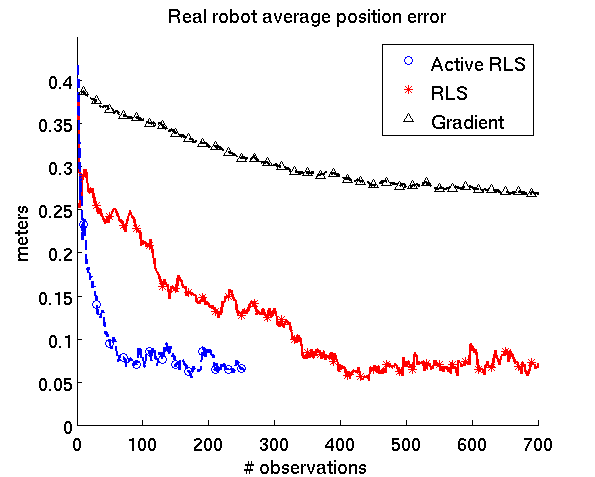}\label{fig:resbaltazar:l}}
\subfigure[]{\includegraphics[width=0.6\columnwidth]{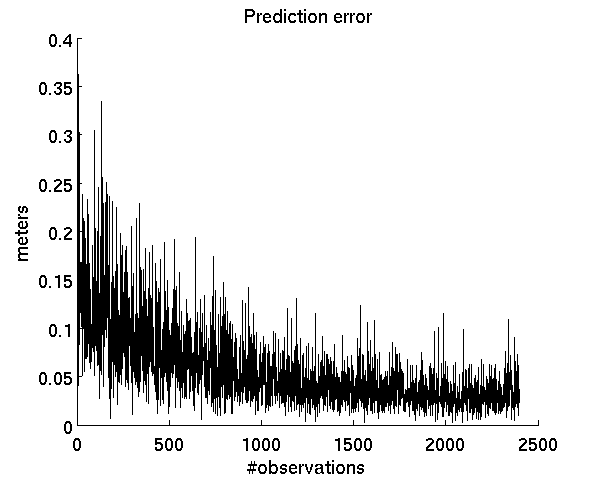} \label{fig:resbaltazar:pred}}
	  \caption{Convergence of the body schema learning for Baltazar with $6$ dof. (a) mean error in the relative orientations of the joint angles, (b) mean
error in the location of the joint angle and (c) prediction error during the learning. Each figure shows the results for random exploration using the RLS, the active approach and an stochastic gradient method.}
	\label{fig:resbaltazar}
\end{figure*}

\begin{figure*}
\centering
\includegraphics[width=0.5\columnwidth]{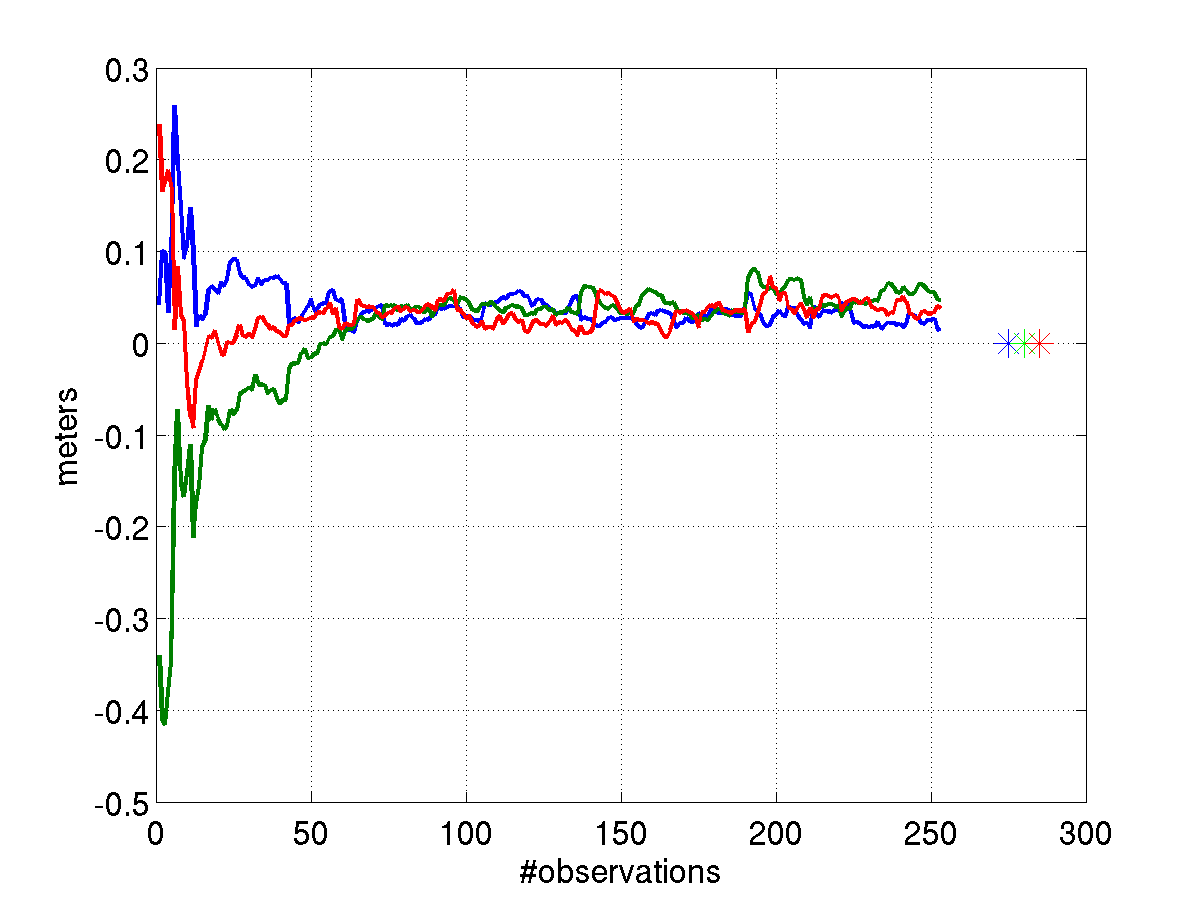} 
\includegraphics[width=0.5\columnwidth]{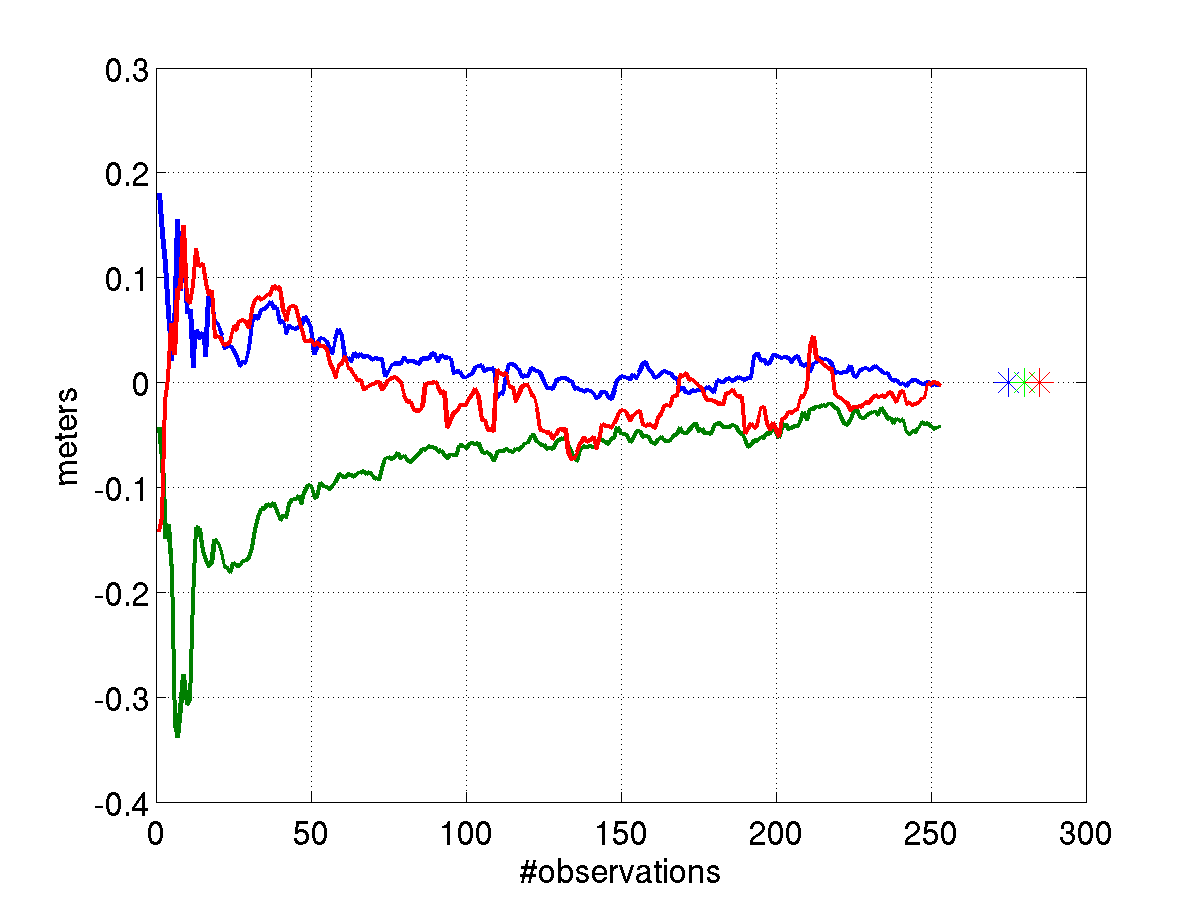} 
\includegraphics[width=0.5\columnwidth]{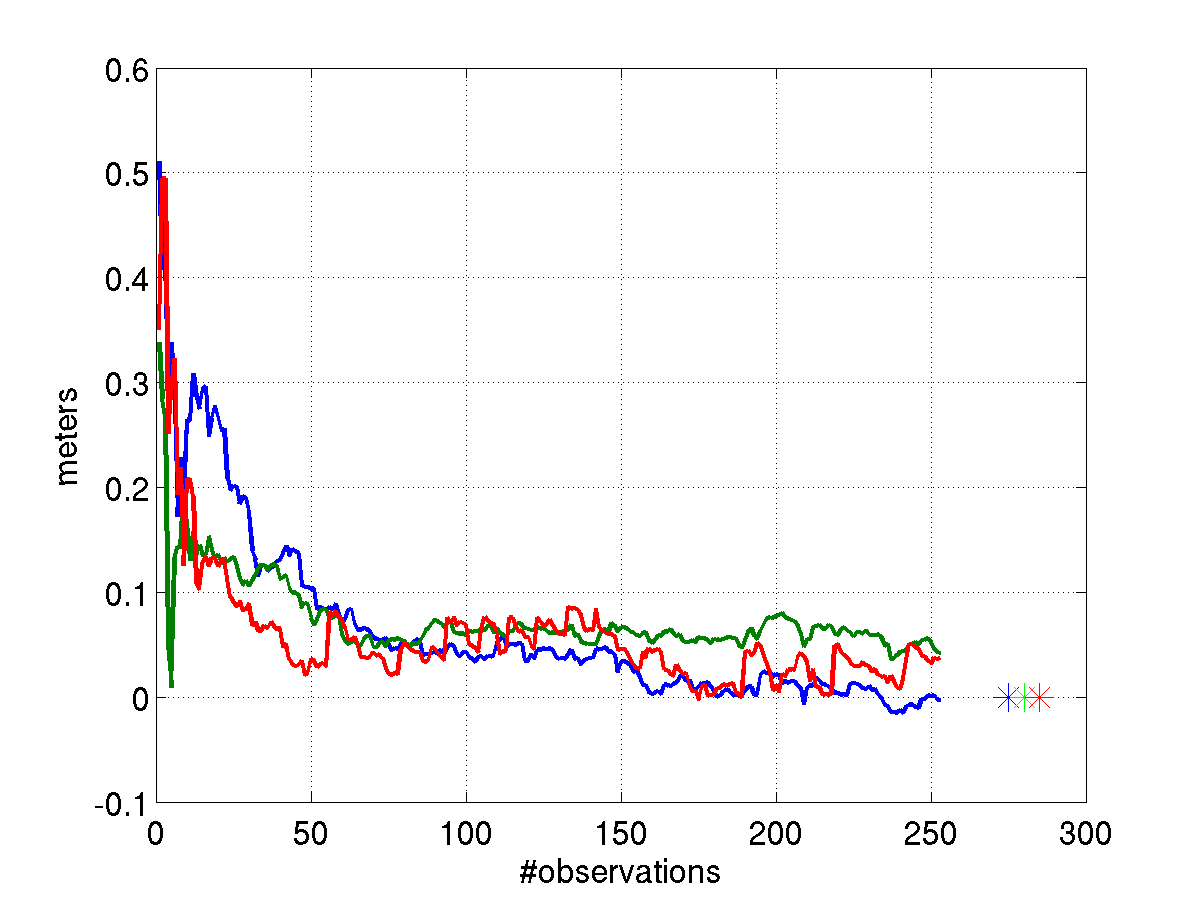} 
\includegraphics[width=0.5\columnwidth]{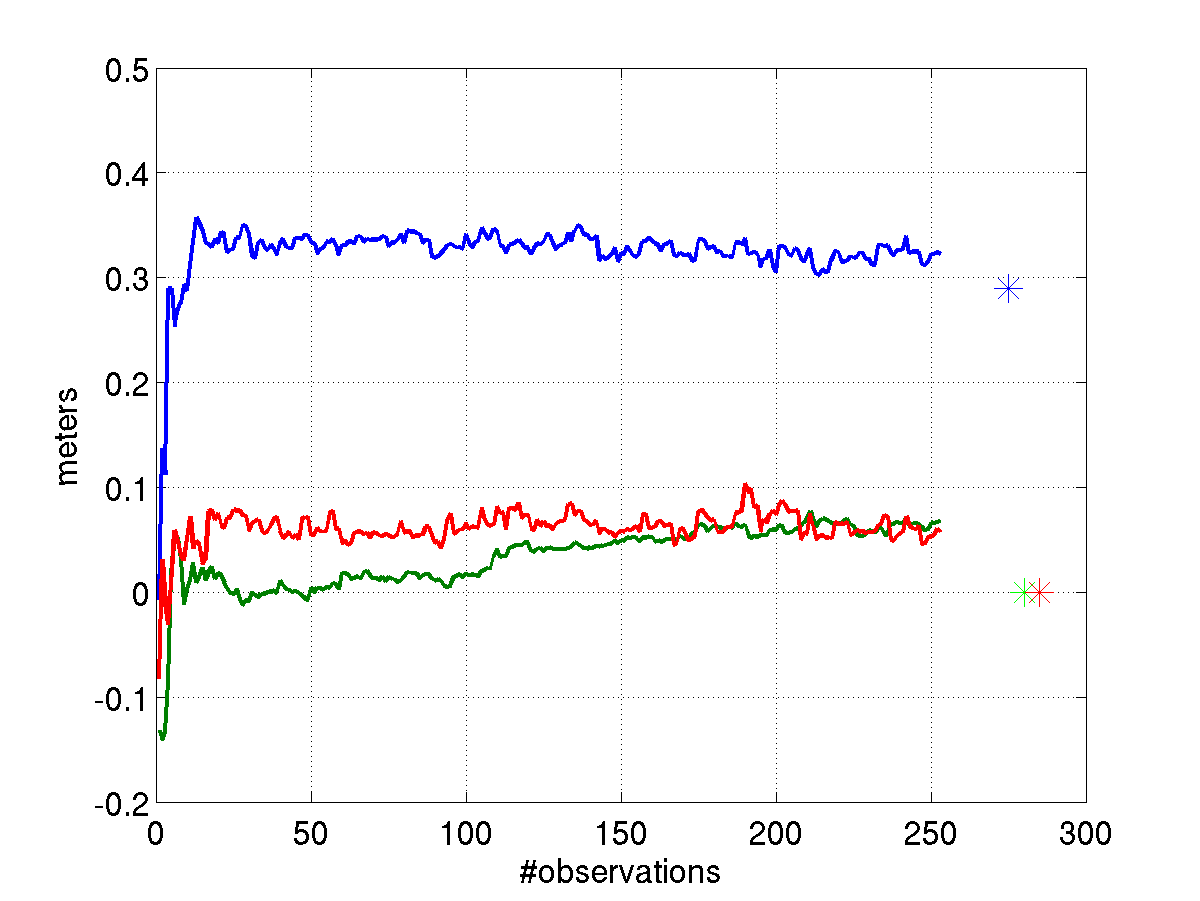} \\
\includegraphics[width=0.5\columnwidth]{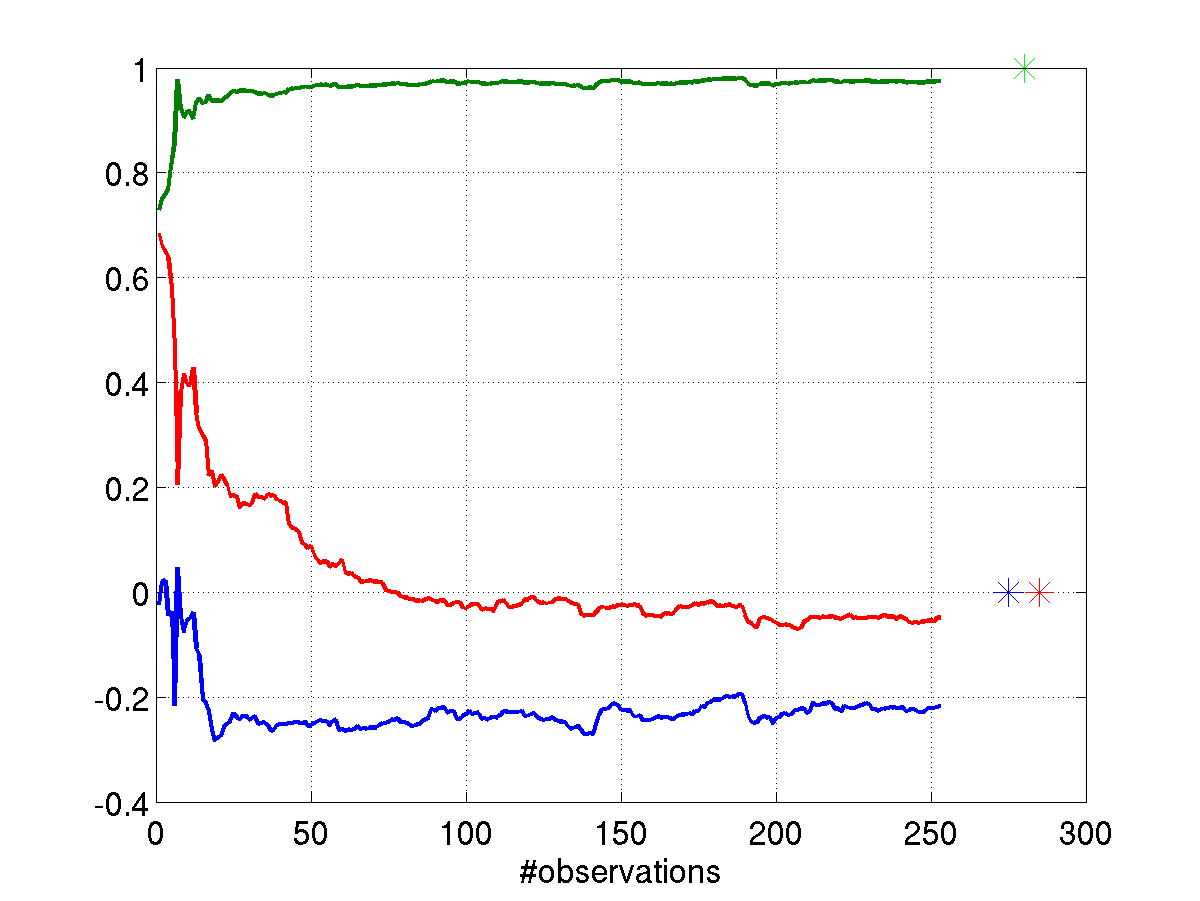} 
\includegraphics[width=0.5\columnwidth]{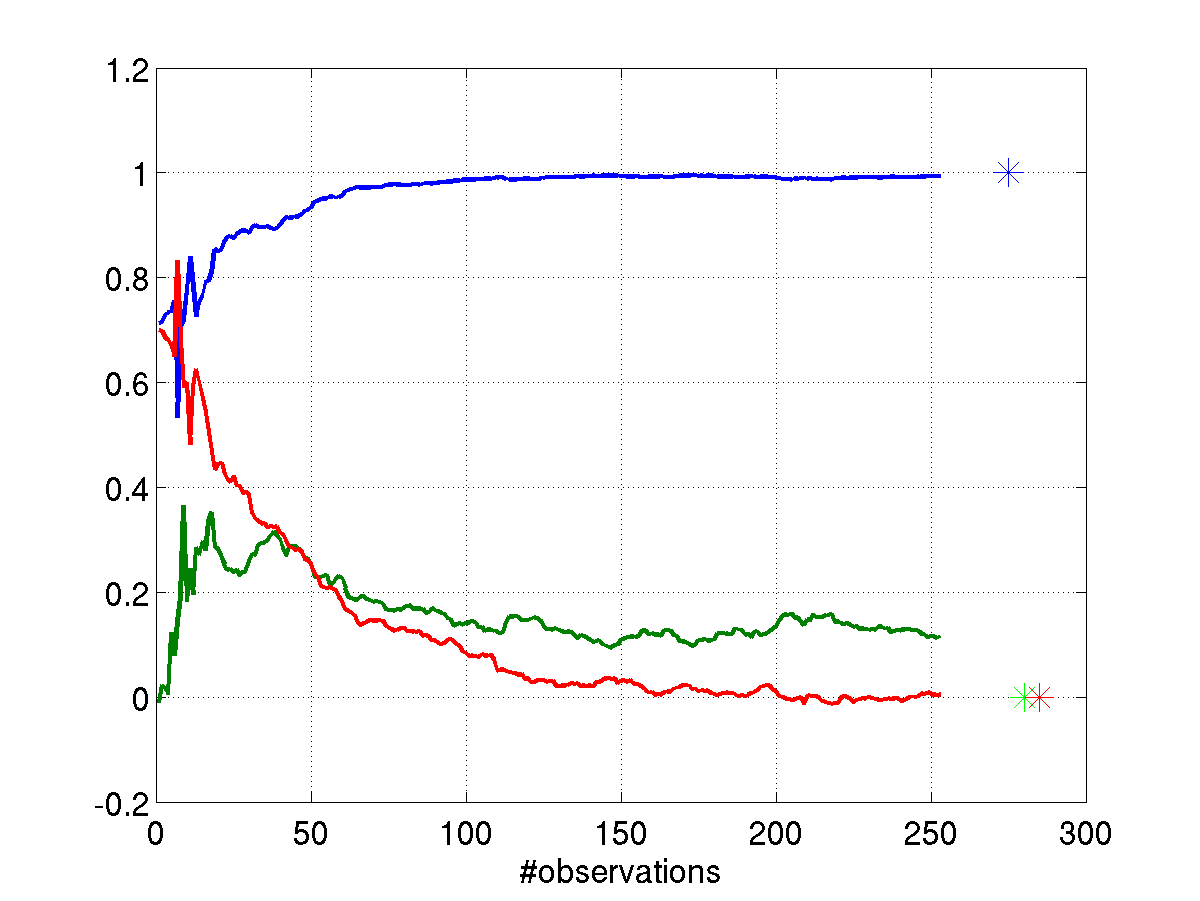} 
\includegraphics[width=0.5\columnwidth]{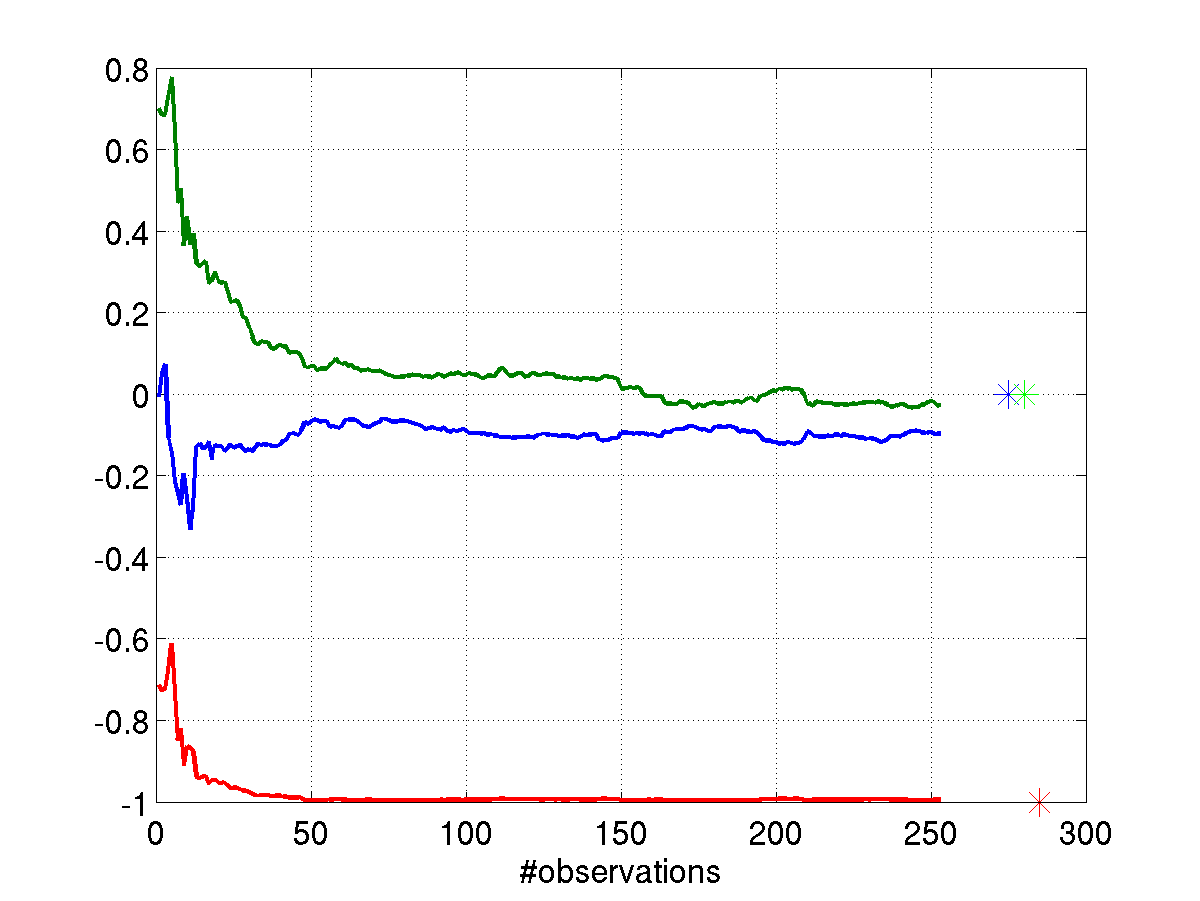} 
\includegraphics[width=0.5\columnwidth]{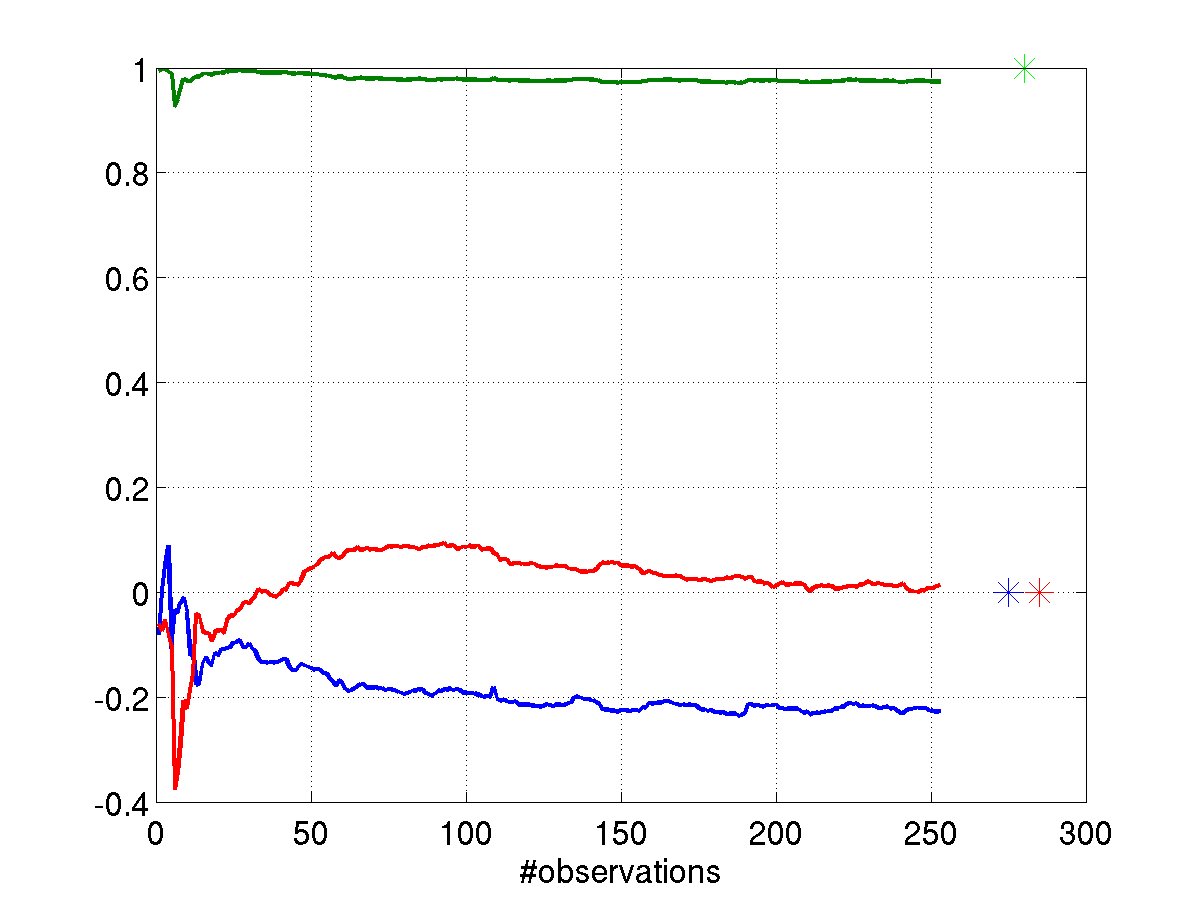} \\
\caption{Joint convergence for a real robot active learning run. (a-d) show the convergence for the joint position (top) and for the direction (bottom). The real values obtained from the original design are shown as asterisks.}
	\label{fig:realconvergence}
\end{figure*}

\subsection{Experiments with the Robot}
\label{sec:experiments}

We next report results on the real robot. From the analysis done in the previous section, we constrained the learning only to the first four joints. This is because our vision system limited the exploration volume which makes the last two degrees of freedom difficult to estimate due to observability issues. As commented before, the signal-noise ratio of the wrist movements is very small or close to 1.

To validate the RLS and the gradient method we acquired a dataset consisting of
$2000$ points taken from random joint positions were each joint had a maximum
allowed excursion of $40 \deg$, which represents the maximum volume for our head
even using servoing. For this experiment, the noise model was tuned to
$\sigma^2_R = 0.01$, a conservative value given our measured errors. In
addition, we added an artificial state noise $\Q=\sigma_Q^2{\bf I}$ with
$\sigma^2_Q=0.005$, that also improved the behavior of the RLS. Figure
\ref{fig:resbaltazar} shows the results for one of the runs.  

The RLS converged in about $600$ observations. The gradient, on the other hand, is still far from the solution, and even after the $2000$ samples it does not converges. We tried several learning rates with similar results. However, the gradient method is able to reduce the prediction error in about $1500$ iterations (see Fig. \ref{fig:resbaltazar:pred}. This is in line with recent results reported on body schema learning \cite{micha08bodyschema} for a similar complex model. 

Finally, the active learning strategy achieved the best results among the three methods. Figures \ref{fig:resbaltazar:ang} and \ref{fig:resbaltazar:l} show that the method was able to converge in less than $80$ iterations. As in the previous simulations, the error was smaller than for the random exploration approach in terms of axis orientation accuracy. Figure \ref{fig:realconvergence} shows the convergence of each individual joint to its true values. 

It is important to note that our solution is partly local and depends on the initialization. The initial estimate had an error of $45\deg$ per orientation angle for all the joints and average error of $0.4 m$ per dimension in the axis locations. When doubling the initialization error, the active learning algorithm required $250$ observations while the RLS increased to $1400$.

\section{Conclusions}
\label{sec:conclusions}

We presented an algorithm to actively learn the body schema of generic open kinematic chains. This is done based only on $3D$ measurements of the end effector of the robot taken for different robot configurations.  Our methods' contributions are: (i) the use of a \emph{Recursive Least Squares} approach to estimate more efficiently the geometric parameters and (ii) a criteria to select online the most informative robot configuration to reduce the uncertainty in the parameters. This selection is based on the expected posterior distribution that is computed simulating the observation from different configurations of the robot. The optimal configuration is the one that minimizes the trace of the covariance of the expected posterior distribution. The optimal value is found using DIRECT, an efficient global optimization method that can be run in real-time.

We evaluated the method to learn kinematic chains, both in simulation and in a real robotic platform. The experimental results show that our active algorithm is able to learn kinematic chains with far less observations than previous methods. For the real robot experiments, our approach requires less than 10\% of the observations required by other methods. The results also suggest that, by selecting the more informative observations, the active strategy usually achieves a better learning performance.

In the future we want to validate our system in higher dof robots and with the use of other sensors (e.g., magnetic trackers or tactile sensors). The use of more accurate sensor can increase the signal-noise ratio of all the joints, allowing the estimation of the full kinematic chain in the real robot. Also we want to reduce the locality in the linearized estimation of RLS. To do this we will need to use more complex distributions and estimation algorithms. We are currently studying the use of the unscented transformation and gaussian mixtures.

\section*{Acknowledgment}
This work was supported in part by the FCT Programa Operacional Sociedade de Informa{\c c}{\~a}o (POSC) in the frame
of QCA III, PTDC/EEA-ACR/70174/2006 project and SFRH/BPD/48857/2008 grant; and in part by the EU Project Handle (EU-FP7-ICT-231640). 

\bibliographystyle{IEEEtran}
\bibliography{IEEEabrv,../bibs/robots,../bibs/macl,../bibs/development,../bibs/biologykin,../bibs/rl,../bibs/affordance,../bibs/computervision,citruben}

\end{document}